\documentclass[11pt]{article}

\usepackage[letterpaper,margin=1in]{geometry}

\usepackage[T1]{fontenc}
\usepackage[utf8]{inputenc}
\usepackage{microtype}

\usepackage{amsmath}
\usepackage{amssymb}
\usepackage{amsthm}
\usepackage{mathrsfs}   % \mathscr{U} -- the governed update operator

\usepackage{graphicx}
\usepackage{booktabs}
\usepackage{array}
\usepackage{ragged2e}
\usepackage[font=small,labelfont=bf]{caption}
\graphicspath{{figures/}}

\usepackage{authblk}
\usepackage[hidelinks,breaklinks]{hyperref}
\usepackage{url}
\theoremstyle{definition}
\newtheorem{definition}{Definition}
\newtheorem*{decomposition}{Operational Decomposition of Accountability}

\newenvironment{gcsblock}{\begin{quote}\small}{\end{quote}}

\newcommand{\aware}{\mathrm{aware}}
\newcommand{\Eaware}{E^{\aware}_{\le\tau}}          % evidence known by tau
\newcommand{\Aaware}{A^{\aware}_{\le\tau}}          % care actions known by tau
\newcommand{\bel}[1]{b_{\tau}(#1)}                  % belief state
\newcommand{\xbranch}{\tilde{x}^{\,\mathrm{branch}}_{t}}   % simulated state
\newcommand{\Xbranch}{\widetilde{X}^{\,\mathrm{branch}}}   % simulated branch
\newcommand{\Bstate}{B_{\tau}}                      % stored belief-state object
\newcommand{\updop}{\mathscr{U}}                    % governed update operator
\newcommand{\dsep}{\,\ensuremath{\cdot}\,}          % separator inside tables

\title{\textbf{Generated Context versus Governed State:\\
Functional Conditions for Accountable Longitudinal Clinical\\
Reasoning}}

\author[1]{Augusto Bernardo Pissarra\thanks{\texttt{augusto.pissarra@myndware.com}}}
\author[1]{Victor Lorena de Farias Souza\thanks{\texttt{victor.lorena@myndware.com}}}
\affil[1]{MyndwareMed\thanks{MyndwareMed builds governed clinical-state
infrastructure for healthcare organizations. The evidence layer specified in
this line of work (maturity Level~3) operates in production; the belief layer
(Level~4) is in engineering and the full Clinical World Model (Level~5) in
research. \url{https://www.myndwaremed.com}}}

\date{August 2026}

\begin{document}

% Title-block footnotes use symbols rather than numbers.
\renewcommand{\thefootnote}{\fnsymbol{footnote}}
\maketitle
\renewcommand{\thefootnote}{\arabic{footnote}}
\setcounter{footnote}{0}

\begin{abstract}
Large language models (LLMs) have become the dominant interface of clinical
artificial intelligence, yet the interface they expose --- text in, text out,
one context window at a time --- maintains no explicit, persistent, governed
representation of what is currently true about a patient. This paper argues
that longitudinal clinical reasoning is a state-estimation problem under
partial observability, and that the axis on which clinical AI succeeds or fails
is not the fluency of the model reading the record but the \emph{governance} of
the patient state it reasons over. We distinguish generated context from
governed state; separate five objects that clinical AI habitually conflates
(true state, observations, evidence, belief, and simulated state); define a
tiered governance standard against which any clinical AI system can be audited;
and show that an operational definition of accountability decomposes into four
information requirements --- an immutable evidence ledger with awareness-time
versioning, a belief state distinct from accumulated evidence, an
observation-process model, and claim-level causal typing. We are explicit that
this decomposition is analytic rather than a necessity theorem, and that its
value is conceptual hygiene: it converts ``accountable clinical AI'' from a
slogan into an audit instrument. A six-level maturity framework separates what
a system makes governable from what it can compute, locating current
LLM-centric practice at high capability but low maturity. The paper is fully
self-contained: the four research questions the framework poses are stated in
the introduction, and the conclusion records what the paper establishes toward
each; future work develops the buildable core of the architecture and the
research program toward full Clinical World Models. No empirical result is
claimed here.
\end{abstract}

\section{Introduction}
\label{sec:introduction}

For sixty years, clinical information systems have stored documents about
patients rather than the patient's state itself. Paper charts became scanned
charts; scanned charts became structured documents; departmental records became
Clinical Document Architecture (CDA) exchanges and then Fast Healthcare
Interoperability Resources (FHIR) bundles. Each generation made clinical
documents easier to create, exchange, and display --- while the work of
reconstructing ``what is true of this patient right now'' stayed where it had
always been: in the head of whoever reads the chart. Large language models are
the first technology fluent enough to perform that reconstruction on demand,
and their fluency makes it easy to mistake reconstruction-on-demand for
something standard LLM-centric architectures do not inherently maintain: an
explicit, accountable, continuously governed representation of the patient.
That mistake, and its architectural remedy, are the subject of this paper.

In the last three years, clinical artificial intelligence has become nearly
synonymous with LLMs~\cite{singhal2023,thirunavukarasu2023}. Hospitals, payers,
and health-technology vendors now route documentation, summarization, coding
assistance, and parts of decision support through general-purpose models, most
commonly over retrieval-augmented substrates~\cite{lewis2020}. The productivity
gains are real, and this paper does not dispute them. But the interface a
language model exposes carries no guarantee of a persistent notion of what is
currently true about a person's physiology, no versioned record of how that
truth changed, and no native mechanism for distinguishing an observed fact from
an inferred one, or an absence of evidence from evidence of absence. Whatever
latent structure such a model may learn internally, none of it is exposed as an
object that a hospital can query, audit, correct, or govern.

We should be precise about how contrarian our thesis is, and is not. It is not
a claim that language models should be abandoned, nor that neural
representation is the problem --- we concede below that a language model can be
wrapped in governed external memory, and that the real contrast is generated
context versus governed state, not neural versus symbolic. The operative claim
is narrower and, we think, sharper: governance of patient state --- not the
fluency of the model reading it --- is the axis on which accountable
longitudinal clinical reasoning succeeds or fails.

One reframing organizes everything that follows, and we state it here rather
than let it emerge by page eight: clinical reasoning over a longitudinal record
is state estimation under partial observability~\cite{kaelbling1998} --- closer
in kind to robotics, econometrics, or epidemiological modeling than to
summarization --- that happens to be documented primarily in natural language.
The patient's true state is never possessed; a care process observes it
selectively; a system forms and revises beliefs from the evidence that
survives. Once the problem is seen this way, the question is not which model
reads the record best but what artifact plays the role of the estimator's
state --- and the deficit of current practice is that no governed artifact
plays it. Put as one sentence: current clinical AI research is increasingly
learning patient dynamics; this paper asks a different question --- what state
must a clinical AI system maintain so that its longitudinal reasoning can be
reconstructed, corrected, audited, and safely separated from simulation?
Everything in this paper unpacks that question.

\paragraph{Research questions.}
Fluency is now abundant; governed clinical state is not --- and the gap between
the two is what makes the framework empirical rather than rhetorical. Four
research questions follow from it, stated here so the reader knows from the
outset what the paper is trying to establish; the conclusion records what this
paper achieves toward each, and what remains to be answered next.
\textbf{RQ1} --- does an architecture that maintains persistent, governed
patient state answer questions about a patient's current and historical status
more consistently, and with better-supported provenance, than long-context or
retrieval baselines?
\textbf{RQ2} --- do dynamics over a governed belief state derived from governed
temporal evidence predict multi-horizon clinical outcomes better than
next-event sequence modeling of the raw record?
\textbf{RQ3} --- does a deterministic verifier materially reduce clinically
impossible outputs without materially reducing sensitivity?
\textbf{RQ4} --- does modeling the process that generates clinical observations
recover a material fraction of the performance lost when models transport
across institutions?

\paragraph{Contributions.}
This paper makes three contributions. First, a precise statement of the deficit
of LLM-centric clinical systems: not representation in general, but concrete
governance properties of state, organized into a tiered audit standard
(Section~\ref{sec:generated-vs-governed}). Second, a vocabulary that keeps
separate the five objects clinical AI habitually conflates, together with an
epistemic-type discipline for assertions and absences
(Section~\ref{sec:five-objects}). Third, an operational definition of
accountability and its decomposition into four information requirements, with
an honest account of the decomposition's logical character, plus a six-level
maturity framework that separates representation maturity from computational
capability (Sections~\ref{sec:accountability}--\ref{sec:maturity}). This paper
is self-contained: it is a conceptual and architectural contribution, complete
on its own terms, and no companion reading is required to evaluate it. Future
work develops the buildable core --- the full specification of the evidence
ledger, the operational belief state, and the governed update operator, with
its implementability at population scale --- and the research program toward
full Clinical World Models: the observation-process model, causal
qualification, action-conditioned dynamics, and quarantined simulation.
Everything in this paper is specification and argument; no empirical or
clinical result is claimed.

\section{Background: where clinical AI stands}
\label{sec:background}

General-purpose LLMs encode substantial medical knowledge ---
instruction-tuned models approach expert-level performance on
licensing-exam-style question answering~\cite{singhal2023,nori2023}, with
applications spanning documentation, triage, and decision
support~\cite{thirunavukarasu2023} --- and models trained on next-token
prediction demonstrably learn internal representations beyond surface
statistics. Nothing in this paper depends on denying that. What benchmark
performance does not establish is that a model answering medical questions well
thereby maintains a consistent, auditable representation of a specific patient
over years of documentation: a high exam score is a statement about a model's
marginal distribution over medical text, not about its ability to track, in a
form anyone can inspect, which of two conflicting medication lists is current
and who documented each.

Retrieval-augmented generation (RAG)~\cite{lewis2020} grounds answers in a
specific chart rather than the training distribution, and it is the substrate
many clinical AI products are built on; agent frameworks chain calls and add
tools and memory. But retrieval is a search operation over a document store,
not state estimation over a model of the patient, and each agent in a chain
still reasons over a transient context rather than a shared, governed state
object --- chaining language models does not, by itself, produce a state
machine, although one can build a governed state machine and let language
models operate against it. The agentic frontier is, in fact, already moving in
this direction at the knowledge layer: Agents-K1~\cite{cao2026} replaces flat
retrieval with an agent-native knowledge substrate --- scientific corpora
transformed into structured graphs that preserve entities, claims, evidence,
and provenance for multi-hop reasoning. That movement supports this paper's
thesis while marking exactly where it stops short: a claims-and-evidence graph
over documents governs what the literature \emph{says}; it is not a belief
state about an evolving individual patient, and it carries none of the
multitemporal, correction, or simulation discipline that longitudinal clinical
accountability requires. Clinical decision support, meanwhile, sets an
instructive credibility bar from an earlier era: a rule-based alert can be
traced, line by line, to the guideline that produced it, and across regulatory
regimes the ability to explain a recommendation materially eases deployment.
Purely generative systems do not clear this bar by default --- a point this
paper develops as a governance property, not a capability claim.

\section{Generated context versus governed state}
\label{sec:generated-vs-governed}

An autoregressive language model is trained to estimate
\begin{equation}
P(t_i \mid t_1, t_2, \ldots, t_{i-1}).
\label{eq:autoregressive}
\end{equation}
Two inferences are commonly drawn from equation~\eqref{eq:autoregressive}, and
only one is valid. The invalid inference is that such a model contains no
representation of state, transition, or persistence --- the training objective
constrains the interface, not the internal solution. The valid inference
concerns the interface itself: everything the model knows about a specific
patient at inference time enters through a transient context window and leaves
as generated text. Whatever latent patient representation forms inside the
forward pass has no identity that survives the call, no address at which a
second system could query it, no version history, no semantics an auditor could
check, and no update discipline distinguishing a correction from a
contradiction. \emph{The representation may exist; the object does not.}
Accountable longitudinal clinical reasoning requires the object.

\begin{figure}[htbp]
  \centering
  \includegraphics[width=0.98\textwidth]{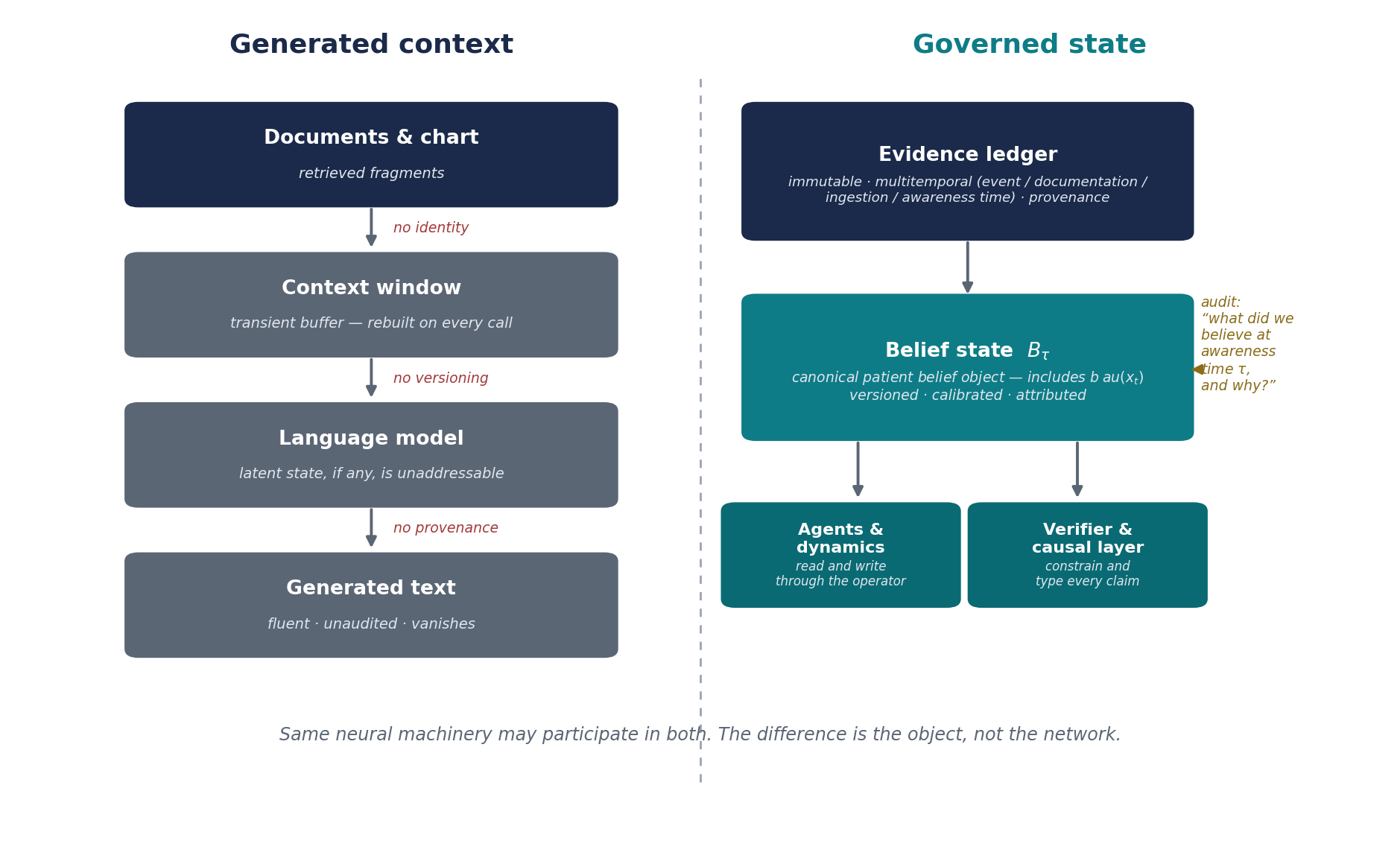}
  \caption{Generated context versus governed state. Left: knowledge of the
  patient is reconstituted into a transient buffer on every call and vanishes
  with it. Right: one canonical, versioned belief state --- fed by an immutable
  evidence ledger --- is the object every component reads from and writes
  through, and the object an auditor queries.}
  \label{fig:regimes}
\end{figure}

The claim in its strongest defensible form: a standard autoregressive
language-model architecture does not provide an externally addressable,
persistent, versioned, and clinically governed patient-state object whose
invariants can be independently verified. Figure~\ref{fig:regimes} shows the
two regimes; the same neural machinery may participate in both, which is
exactly why the difference must be located in the object, not the network. The
properties the governed regime guarantees are concrete engineering
requirements: \emph{identity} (one canonical state object per patient, stably
addressable); \emph{persistence} (survival across calls, sessions, encounters,
years, by construction); \emph{explicit semantics} (ontology bindings, units,
value sets~\cite{snomed,loinc,owl2}); \emph{update governance} (writes through a
defined operator, with rules for who may assert, correct, or supersede);
\emph{provenance} (every element traceable to the evidence that produced it);
\emph{inspectability}; \emph{deterministic constraints}; \emph{calibration};
\emph{rollback and correction}; and \emph{historical query} (``what did we
believe at awareness time~$\tau$, about clinical time~$t$, and why?'').

We treat these properties as a normative engineering standard, not as necessary
and sufficient conditions, and we organize them --- together with the
additional requirements each higher tier introduces --- into tiers matched to
stakes; Definition~\ref{def:gcs} states the tiers, and Table~\ref{tab:tiers}
normalizes their requirements.

\begin{gcsblock}
\begin{definition}[Governed Clinical State, tiered]
\label{def:gcs}
\emph{Core} governed state requires identity, persistence, explicit semantics,
provenance, update governance, and historical query --- the minimum for any
representation other components treat as authoritative about a patient.
\emph{Safety-critical} governed state additionally requires deterministic
constraint validation, correction propagation, and explicit uncertainty
representation --- required wherever state feeds clinical decisions.
\emph{Predictive} governed state additionally requires outcome calibration,
model-version binding, and monitored validity under distribution shift ---
required wherever state includes forecasts or simulations. The definition is
architecture-neutral: it does not require that the representation be
\emph{symbolic} (structure held in explicit, human-readable units with declared
semantics --- graphs, rules, ontology-bound assertions), \emph{neural}
(structure held in learned continuous parameters and activations), or
\emph{hybrid} (a composition of both, with defined interfaces between them) ---
only that the tier's properties hold and can be independently audited. We use
these three terms in exactly this sense throughout.
\end{definition}
\end{gcsblock}

\begin{table}[htbp]
  \centering
  \small
  \begin{tabular}{@{}l>{\RaggedRight}p{0.70\textwidth}@{}}
    \toprule
    \textbf{Tier} & \textbf{Requirements} \\
    \midrule
    Core & identity \dsep persistence \dsep explicit semantics \dsep provenance
           \dsep update governance \dsep historical query (served by
           deterministic replay) \\[2pt]
    Safety-critical & core $+$ explicit uncertainty typing \dsep deterministic
           constraint validation \dsep correction propagation \\[2pt]
    Predictive & safety-critical $+$ outcome calibration \dsep model-version
           binding \dsep monitored validity under distribution shift \\
    \bottomrule
  \end{tabular}
  \caption{The tiers of Definition~\ref{def:gcs} (Governed Clinical State),
  normalized. Each tier inherits the tier below it; a system is audited at the
  tier matching what its state is used for.}
  \label{tab:tiers}
\end{table}

Note what this definition does not say. It does not say a language model cannot
participate in maintaining governed state --- on the contrary, language models
are essential at the perception boundary, reading and drafting the documents
the ledger ingests. It does not say symbolic systems automatically qualify: a
knowledge graph with silent overwrites and no provenance fails the core tier as
surely as a context window does. Persistence and auditability are systems
properties, not gifts of symbolic representation --- a neural system can be
wrapped in event-sourced external memory; symbolic structure earns its place
for specific reasons (checkable semantics, human inspectability, ontology
binding), not because symbols are where persistence lives. And the definition
converts this paper's thesis into an audit instrument: for any proposed
clinical AI system, ask which tier its patient representation reaches, and
demand evidence property by property. Table~\ref{tab:classes} previews the
audit against architectural classes rather than products; the properties it
interrogates are exactly the ones this section defined.

\begin{table}[htbp]
  \centering
  \small
  \begin{tabular}{@{}lccccc@{}}
    \toprule
    \textbf{Architectural class} & \textbf{Persistent} & \textbf{Governed} &
    \textbf{Replay} & \textbf{Belief $\neq$ evid.} & \textbf{Quar.\ sim.} \\
    \midrule
    Electronic health record (EHR) / document store~\cite{fhir}
      & yes & partial & no & no & no \\
    RAG / long-context LLM~\cite{lewis2020}
      & no & no & no & no & no \\
    LLM agent $+$ memory~\cite{wang2024agents,cao2026}
      & partial & no & no & no & no \\
    Knowledge graph (typical)~\cite{hogan2021}
      & yes & partial & no & no & no \\
    Learned world model~\cite{hafner2023,mu2026,adam2026}
      & partial & no & no & no & partial \\
    Governed clinical state (this paper)
      & yes & yes & yes & yes & yes \\
    \bottomrule
  \end{tabular}
  \caption{Architectural classes, not products, against the governance
  properties of Definition~\ref{def:gcs} (Governed Clinical State). The five
  columns compress the audit for readability rather than replacing it:
  Persistent summarizes identity and persistence; Governed summarizes explicit
  semantics, update governance, provenance, deterministic constraints, and
  calibration; Replay summarizes historical query and rollback/correction; the
  last two columns audit the object separations of
  Section~\ref{sec:five-objects} (belief distinct from evidence; quarantined
  simulation), which the tier requirements presuppose. Entries indicate
  properties guaranteed by the architectural class as such, not properties a
  particular implementation could add; ``partial'' marks properties
  implementations may add without a class-level guarantee; the final row states
  what this paper requires, not what any deployed system has demonstrated ---
  the audit questions, not the grades, are the contribution.}
  \label{tab:classes}
\end{table}

\section{Five objects that must not be conflated}
\label{sec:five-objects}

\begin{figure}[htbp]
  \centering
  \includegraphics[width=0.90\textwidth]{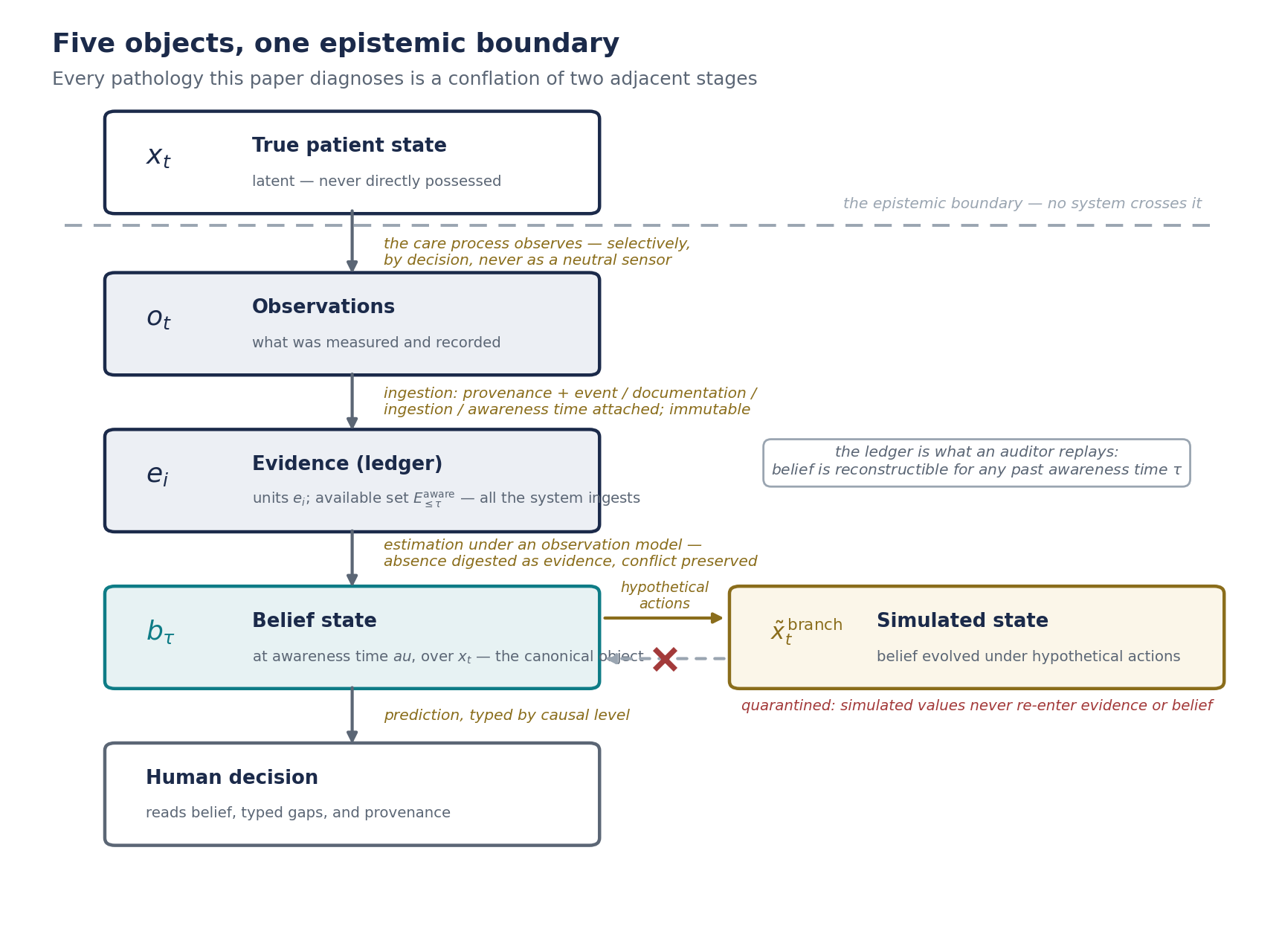}
  \caption{The five objects and the epistemic boundary. The true state $x_t$
  sits above a boundary no system crosses; observations $o_t$, evidence units
  $e_i$, and belief $\bel{x_t}$ are successive constructions below it, and
  simulated state $\xbranch$ lives in a quarantined branch whose values never
  re-enter evidence or belief. The pathologies of LLM-centric practice are
  conflations of adjacent stages --- most commonly, treating evidence as if it
  were belief.}
  \label{fig:five-objects}
\end{figure}

Before the formalism, the picture. A patient exists; a care process looks at
the patient, selectively; what it records becomes evidence; from evidence the
system forms a belief about the patient; from belief it predicts, and in
quarantined branches it simulates; a human decides. Every arrow in that chain
loses, shapes, or creates information (Figure~\ref{fig:five-objects}). The
vocabulary of clinical AI habitually collapses this entire chain into the
single word ``state,'' and that collapse is not a terminological quibble: each
of the stages is a different mathematical object, with different update rules
and different failure modes, and every pathology this paper diagnoses is a
conflation of two \emph{neighboring} stages of this chain: treating
observations as the truth ($o_t$ mistaken for $x_t$), treating extracted
assertions as established facts ($e_i$ mistaken for $o_t$), treating the
accumulated record as what the system believes ($\Eaware$ mistaken for
$b_\tau$ --- the most common and most damaging), or letting simulated values
leak back into the record ($\xbranch$ contaminating ledger or belief).

This paper therefore assigns each stage its own symbol and keeps all five
separate:

\begin{center}
\begin{tabular}{@{}ll@{}}
  $x_t$ & true (latent) patient state at time $t$ --- never directly possessed \\[2pt]
  $o_t$ & observations available at time $t$ --- produced by the care process \\[2pt]
  $e_i$ & a unit of evidence,
          $(o_i, t^{\mathrm{event}}_i, t^{\mathrm{doc}}_i,
            t^{\mathrm{ingest}}_i, \tau^{\aware}_i, p_i)$ ---
          what the system ingests \\[2pt]
  $\bel{x_t}$ & belief at awareness time $\tau$ about clinical time $t$ ---
          the canonical object \\[2pt]
  $\xbranch$ & simulated state under a hypothetical --- quarantined from
          governed belief \\
\end{tabular}
\end{center}

The true state $x_t$ is the patient's actual physiological and clinical
condition; no system --- and no clinician --- possesses it. Observations $o_t$
are what the care process happened to measure and record --- always the product
of decisions, never a neutral sample. Evidence comes in immutable units
$e_i = (o_i, t^{\mathrm{event}}_i, t^{\mathrm{doc}}_i, t^{\mathrm{ingest}}_i,
\tau^{\aware}_i, p_i)$: an observation wrapped with provenance $p_i$ and four
time stamps --- event, documentation, ingestion, and awareness time, the last
being when the system's belief first reflected it. Corrections create new
evidence linked to old, never edits in place. The evidence available at
awareness time $\tau$ is the set
$\Eaware = \{\, e_i : \tau^{\aware}_i \le \tau \,\}$ --- exactly the
conditioning set of equation~\eqref{eq:belief} --- so the chain runs
$x_t \to o_i \to e_i \to \Eaware \to \bel{x_t}$. The belief state requires two
clocks, and conflating them would undo the very multitemporal distinction this
paper claims as a contribution. Let $t$ denote \emph{clinical time} --- the
time at which the patient's latent state held --- and $\tau$ denote
\emph{awareness time} --- the time at which the system's knowledge is being
evaluated. The belief state is then the system's probability distribution over
the true state at clinical time $t$, given everything the system was aware of
by $\tau$:
\begin{equation}
\bel{x_t} = P\!\left(X_t = x \mid \Eaware,\ \Aaware\right),
\label{eq:belief}
\end{equation}
where $\Eaware$ is all evidence with awareness time at most $\tau$ and
$\Aaware$ the care actions --- prescriptions, procedures, orders --- known to
the system by $\tau$. The two indices are what make replay a mathematical
statement rather than a slogan: $b_{\tau_1}(x_t)$ versus $b_{\tau_2}(x_t)$ asks
what did the system believe at awareness time $\tau_1$ about the patient's
condition at clinical time $t$, compared with what it believed later ---
exactly the comparison a delayed discharge summary forces, where an event of
January ($t$) enters awareness only in February ($\tau$), and exactly the query
an auditor asks after a correction. Where the two clocks coincide we write
$b_t$ as shorthand, but the object is always $\bel{x_t}$. We reserve $\Bstate$
for the complete stored belief-state object --- the software artifact, one per
patient --- of which each $\bel{x_t}$ is a component posterior about clinical
state at one time; the composition claim of Section~\ref{sec:positioning} is
stated over $\Bstate$. How $\bel{x_t}$ is parameterized computationally ---
factored parametric distributions, probabilistic-logic assertions over
ontology-bound variables, neural latent state with declared decoders, or a
hybrid --- is deliberately left open at this level: the requirements below
constrain any choice, and an operational minimum profile is future work. The
belief state is required to be versioned, calibrated where its components admit
calibration, attributed, and honest about ignorance --- unmeasured represented
as unmeasured, not imputed to normal --- and it may become multimodal when
evidence genuinely supports incompatible alternatives. Simulated state
$\xbranch$, finally, is belief evolved under hypothetical actions, and its role
is prospective: it is the object that answers ``what if?'' --- comparing
candidate treatment plans, projecting a trajectory under an alternative
medication, stress-testing a discharge decision --- before any action is taken
on the real patient. That usefulness is exactly why its quarantine matters:
branches can be created, compared, and discarded freely only because they live
where they cannot contaminate the record of what was actually observed, and
simulated values never re-enter the ledger or the belief. Systems that collapse
the first four objects ($x_t$, $o_t$, $e_i$, $b_\tau$) into one --- most
commonly by defining state as the accumulated record --- become simulators of
documentation rather than estimators of the patient, however well they predict.

A worked contrast makes the belief object concrete. Two active medication lists
survive ingestion: the hospital discharge list includes apixaban; the
primary-care list, updated later by a clinic that may not have known about the
admission, does not. A record-as-state system answers ``is this patient
anticoagulated?'' with whichever fragment retrieval happens to surface. The
belief state instead answers with an explicitly uncertain --- possibly
multimodal --- representation, preserving the evidence chains that support each
alternative; how probability mass is assigned across alternatives is a
calibration question this paper deliberately leaves to future work. The
uncertainty is the answer; a system that hides it has not resolved the
conflict, only concealed it.

\subsection{Epistemic types, and the taxonomy of absence}
\label{sec:epistemic-types}

Governed state requires that every assertion carry its epistemic type --- the
answer to ``how do we know this?'' --- as a machine-checkable attribute rather
than a nuance of phrasing. Two orthogonal attributes must not be merged. The
\emph{epistemic origin} answers how the assertion was produced: observed,
reported, inferred, predicted, simulated, guideline-derived, or unknown. The
\emph{assertion status} records its standing relative to other evidence:
supported, contradicted, superseded, unresolved-conflict, or retracted --- the
attribute through which the managed consistency of the next subsection
operates. Two rules make the taxonomy load-bearing. First, type is inherited
and never laundered: an inference built on a reported value is at most as
strong as the report, and a prediction can never be silently re-typed as an
observation. Second, absence is typed with the same care as presence:
\[
  \text{unknown} \neq \text{negative} \neq \text{not measured}
  \neq \text{not documented} \neq \text{unavailable.}
\]
The canonical illustration is the allergy question. ``No allergies
documented'' (nothing was asked), ``patient reports no allergies'' (a reported
negative), ``allergy testing negative'' (an observed negative), and ``allergy
list not accessible from this source'' (an availability fact --- which is why
unavailable is a first-class member of the taxonomy: absence at this system is
not absence in the world) authorize four different clinical actions --- and a
system that renders all four as the reassuring phrase ``no known allergies''
has erased a patient-safety distinction. In text, these distinctions survive
only if a writer chose careful words and a reader parses them; in governed
state, they are types, and a verifier can refuse an action whose preconditions
demand an observed negative where only an undocumented absence exists.

\subsection{Managed consistency, not guaranteed consistency}
\label{sec:managed-consistency}

It is tempting to demand that a clinical representation be guaranteed free of
contradictions. The demand is misguided, because clinical evidence is
legitimately contradictory: clinicians disagree; assays disagree; a diagnosis
is provisional and then revised; documentation lags reality. A system that
guarantees a contradiction-free representation can do so only by forcing
conflicting evidence into a single truth --- precisely the silent merging that
destroys auditability. What we require instead is \emph{managed consistency}:
contradictions are detected mechanically; contradictory evidence is preserved,
never overwritten; incompatible assertions are never silently merged; belief
estimation consumes the disagreement as evidence structure --- widening or
splitting $\bel{x_t}$ rather than pretending to a resolution the evidence does
not license; and conflicts remain visibly open until evidence, or a human,
closes them. A diagnosis copied forward for two years after the infiltrate
resolved is not deleted but superseded --- and an auditor can still see exactly
how long the stale assertion survived and which report ended it.

\section{One question, asked of both architectures}
\label{sec:one-question}

``Can this patient safely receive iodinated contrast today?'' looks like a
retrieval problem. It is, in miniature, everything this paper formalizes: it
requires current renal function (not the most recently documented creatinine);
the status of metformin therapy, where relevant to medication management around
contrast administration (an order is care-process state, not a sentence in a
note); prior contrast reactions (where the difference between ``no reaction
documented'' and ``no reaction'' can be the entire answer); and knowledge of
what the record is missing (prior imaging happened at another institution).
Table~\ref{tab:contrast} shows how the two regimes of
Figure~\ref{fig:regimes} fare.

\begin{table}[htbp]
  \centering
  \small
  \begin{tabular}{@{}>{\RaggedRight}p{0.16\textwidth}
                    >{\RaggedRight}p{0.34\textwidth}
                    >{\RaggedRight}p{0.40\textwidth}@{}}
    \toprule
    \textbf{Component} & \textbf{Generated context (RAG / long-context)} &
    \textbf{Governed state} \\
    \midrule
    Renal function &
      Treats a 14-month-old creatinine as the patient's current renal status
      --- fluent, confident, stale &
      Belief over renal function widened by 14 months without measurement:
      ``unknown; last known value $x$; a point-of-care creatinine would resolve
      it'' \\[4pt]
    Metformin &
      Whichever fragment ranks higher wins; the conflict itself is invisible &
      Contradicted-open: both evidence chains preserved, belief explicitly
      uncertain, conflict flagged for human closure \\[4pt]
    Prior reaction &
      Renders absence as ``no known allergies'' &
      Typed unknown/unavailable, not negative: prior contrast records exist at
      a source unavailable to this system \\[4pt]
    The gaps themselves &
      No mechanism: retrieval returns documents, not the absence of documents &
      First-class output: the typed gaps and the measurements that would close
      them \\
    \bottomrule
  \end{tabular}
  \caption{One clinical question, component by component. The failure on the
  left is not that the language model reads badly --- it reads perfectly. The
  failure is that four of the five objects of Section~\ref{sec:five-objects} do
  not exist anywhere in that architecture, so no amount of reading can consult
  them.}
  \label{tab:contrast}
\end{table}

Notice the output the governed architecture can represent: not ``yes'' or
``no,'' but a set of typed gaps and the actions that would close them --- a
language model may of course recommend such a measurement; the architectural
distinction is that a governed observation model can represent the missing
measurement itself as a typed information gap and justify the measurement as an
uncertainty-reducing action (the observation-process model, developed in future
work).

\subsection{The contrast in full: a longitudinal case}
\label{sec:longitudinal-case}

\begin{figure}[htbp]
  \centering
  \includegraphics[width=0.99\textwidth]{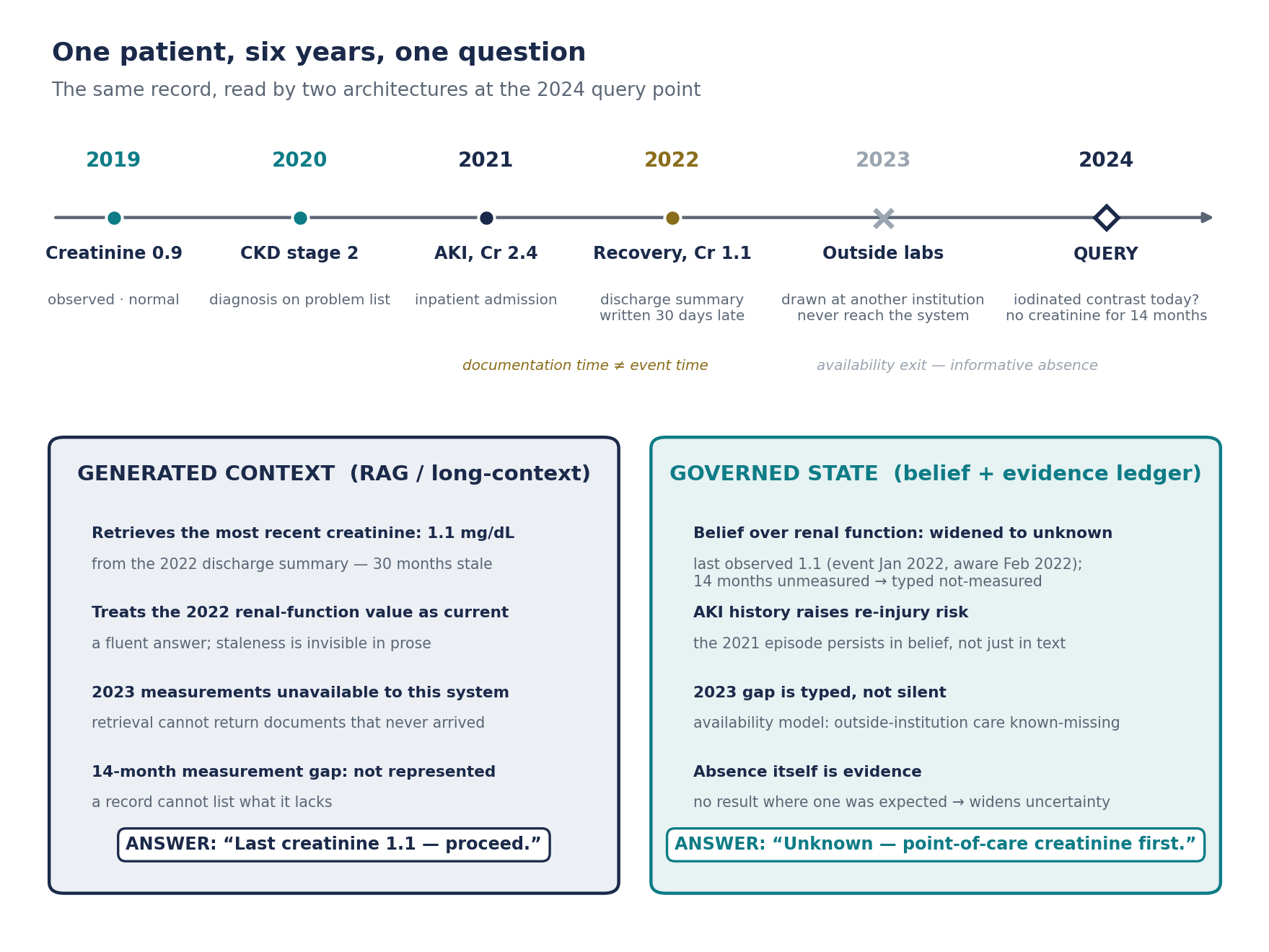}
  \caption{One patient, six years, one query. Top: the timeline, with a
  documentation lag (2022), an availability exit (2023), and an informative
  measurement gap (2023--24). Bottom: the same record read at the same moment
  by the two architectures of Figure~\ref{fig:regimes}. The governed answer is
  not produced by a better reader but by objects the generated-context
  architecture does not possess. The case is an illustrative construction, not
  an executed comparison: the claim is about mechanisms each architecture can
  and cannot express, not measured performance.}
  \label{fig:six-years}
\end{figure}

This subsection develops what the section has just presented in more detailed
and practical form: the same contrast, now with a concrete timeline --- dates,
values, documents, and gaps --- rather than isolated components. The
component-level table above is deliberately synchronic; the failure modes that
motivate this paper are diachronic --- they accumulate over years --- so we
close the argument with a case-based illustration: one realistic six-year
record, read by both architectures at the same query point
(Figure~\ref{fig:six-years}).

The record: a creatinine of 0.9~mg/dL in 2019 (normal); chronic kidney disease
(CKD) stage~2 added to the problem list in 2020; an acute kidney injury (AKI)
with creatinine peaking at 2.4 during a 2021 admission; recovery to 1.1 in
January 2022 --- documented in a discharge summary written thirty days after
the events it describes; laboratory work drawn at an outside institution in
2023 that never reaches this system; and then nothing --- no measurement for
fourteen months --- until the 2024 query: can this patient safely receive
iodinated contrast today?

Each event exercises a different object from Section~\ref{sec:five-objects}.
The 2022 discharge summary makes documentation time diverge from event time: a
system that conditions on it as of January has silently read the future. The
2023 outside labs are an availability failure: the patient's true state was
measured, and the measurement exists --- in a silo this system cannot see ---
so the absence in the record is not absence in the world. And the
fourteen-month gap is itself potentially informative about the observation
process --- but it does not identify its cause: follow-up may not have
occurred, may have occurred outside the observable system, or may have
generated evidence that never reached this one. What governed state owes the
clinician is exactly that typed ambiguity, not a story; converting a missing
observation into one specific inferred explanation would repeat, in reverse,
the overconfidence this paper diagnoses.

A record-as-state, generated-context architecture can produce a fluent but
unjustifiably confident answer. Retrieval surfaces the most recent
creatinine --- 1.1, from the 2022 summary --- and treats it as the patient's
current renal status; the value's staleness is invisible in prose, the 2023
outside measurements are simply unavailable to this system (retrieval cannot
return documents that never arrived), and the measurement gap is not
represented anywhere, because a record cannot list what it lacks. The answer is
``last creatinine 1.1 --- proceed,'' delivered with the confidence of a system
that has read everything it was given. The governed-state architecture instead
represents the unresolved uncertainty explicitly --- and not because its reader
is smarter: the same language model may sit at its perception boundary. The
belief over renal function has widened with every unmeasured month since the
last awareness-timed observation; the 2021 AKI persists in belief, as a risk
modifier, rather than only in a note; the 2023 gap is typed as an availability
failure rather than rendered as reassurance; and the absence of an expected
measurement has itself been digested as evidence. The answer is ``current renal
function unknown; last known 1.1 in January 2022; a point-of-care creatinine
would resolve it'' --- a set of typed gaps and the action that closes them.
Every difference between the two columns of Figure~\ref{fig:six-years} traces
to exactly one cause: the objects of Section~\ref{sec:five-objects} exist in
one architecture and not the other.

\section{An accountability decomposition}
\label{sec:accountability}

\subsection{The demand}
\label{sec:the-demand}

We call a system's longitudinal clinical reasoning \emph{accountable} if, for
any patient-specific temporal conclusion it produces, a third party can:
(R1)~reconstruct the evidence the system was aware of when it produced the
conclusion; (R2)~distinguish what the system observed from what it inferred,
predicted, or simulated --- including distinguishing observed absence from
unmeasured absence; and (R3)~determine what kind of claim, associational or
causal, the conclusion makes. Nothing in this definition mentions architecture;
it is a demand a hospital, a court, or a regulator can state without knowing
how the system works.

\subsection{The decomposition}
\label{sec:the-decomposition}

\begin{gcsblock}
\begin{decomposition}
Given requirements R1--R3, the required information content decomposes into
four functional artifacts: any system whose longitudinal clinical reasoning is
accountable in the sense above maintains, explicitly or in an informationally
equivalent form, four artifacts: (A1)~an immutable evidence ledger with
awareness-time versioning; (A2)~a belief state distinct from accumulated
evidence, with epistemic typing of its contents; (A3)~an observation model ---
an account of how observations came to exist; and (A4)~claim-level causal
typing of every predictive output. We say \emph{artifacts} to keep them
distinct from the five epistemic objects of Section~\ref{sec:five-objects}, to
which they map directly: A1 materializes the evidence units $e_i$ and their
ledger, A2 materializes the belief object $\bel{x_t}$, A3 models the passage
from $x_t$ to $o_t$, and A4 types the predictions derived from $b_\tau$. The
decomposition asserts functional necessity only: an accountable system must
maintain the information content and update discipline of A1--A4, but need not
expose them as four physically separate services or data structures --- belief
and evidence may share storage, an observation model may be folded into a
generative data model, and causal qualification may be enforced at query time,
provided the functional properties hold and can be audited.
\end{decomposition}
\end{gcsblock}

The argument is elimination, one requirement at a time. R1 forces A1: to
reconstruct ``what the system was aware of at awareness time $\tau$,'' the
information distinguishing awareness time from event time must have been
preserved at ingestion --- it cannot be recovered afterward from content alone,
since a discharge summary describing day~2 reads identically whether it arrived
on day~3 or day~30. R2 forces A2: if no functional distinction between evidence
and belief is maintained, observed and inferred are distinguished nowhere, and
the distinction cannot be reconstructed after the fact. R2's second clause
forces A3: the difference between ``tested negative'' and ``never tested''
cannot in general be recovered reliably from the record's content alone; it
lives in the process that generated the record. R3 forces A4: a conclusion's
causal status is a property of how it was derived, and unless attached at
derivation time, an associational estimate and an interventional one are
indistinguishable downstream.

\subsection{What kind of result this is --- and is not}
\label{sec:what-kind-of-result}

We should be candid about the logical character of the argument. Because we
define accountability as R1--R3 and those requirements essentially name the
information the four objects carry, the ``forcing'' is analytic --- a
clarification of what our definition already entails --- rather than a
synthetic discovery about the world. That is a genuine and, we think, useful
contribution: it pins down what accountability must mean operationally and
shows the definition is coherent and decomposable. But it is not a theorem in
the strong sense, and it should not be read as one; its force is conceptual
hygiene. A reviewer who exhibits an accountable system that factors the objects
differently confirms the decomposition rather than refuting it, so long as the
functional properties survive. Its empirical counterpart --- whether
maintaining the four artifacts actually improves clinical reasoning --- is a
separate question, staked as the research questions summarized in the
conclusion.

Two consequences follow. \textbf{Context-window insufficiency:} an architecture
in which patient information exists at reasoning time only as context assembled
for the call --- whatever the scale or quality of the model that reads it ---
maintains none of the four artifacts, and its longitudinal reasoning cannot be
made accountable solely by prompting, retrieval, fine-tuning, or increased
context length: accountability requires persistent governed information outside
the transient context itself --- changing the architecture, not the model.
\textbf{The evaluation shift:} if the decomposition holds, evaluation of
clinical AI intended for longitudinal use must weigh the governance properties
of the patient state it maintains at least as heavily as the quality of the
text it generates. Output-only benchmarks cannot distinguish a system that
maintains the four artifacts from one that merely sounds as if it does. The
audit questions of Definition~\ref{def:gcs} (Governed Clinical State) can.

\section{A maturity framework for clinical representations}
\label{sec:maturity}

Clinical knowledge is, overwhelmingly, encoded in text, and good clinical prose
carries temporality, uncertainty, negation, causal language, hypotheses, and
disagreement. The defensible claim is operational: the structure text carries
is implicit, inconsistently expressed, entangled with authorship, and therefore
not queryable, not validatable, not updatable, and not governable at the
standard of Definition~\ref{def:gcs} (Governed Clinical State). Take one
sentence of ordinary clinical prose: ``Given her worsening renal function, we
held the lisinopril and will recheck creatinine on Monday; if stable, resume at
half dose.'' In one line: an observation trend, a causal attribution, an
executed action with a reason, a scheduled observation, a conditional plan, and
an implicit baseline. A language model, prompted well, extracts most of it,
most of the time. But the sentence's structure becomes usable infrastructure
only when each element lands in governed state with its epistemic type --- and
left as prose, every distinction survives only until the next summarization,
the next copy-forward, the next context-window eviction. The failure mode of
text is not that it cannot express structure; it is that nothing enforces the
structure it expresses.

Before walking any levels, the distinction that makes a maturity framework
defensible at all: two properties of a representation must be kept apart ---
\emph{representation maturity}, what the system makes explicit and governable,
and \emph{computational capability}, what it can infer internally. A multimodal
time-series model can estimate a latent physiological state without ever
building an entity graph; a mechanistic digital twin can host world-model
machinery over tabular variables with no NLP anywhere. The framework's ordering
claim is therefore about governance, not capability, and it does not assert
that systems must be built level by level: a system can compute at level $k$
while only being accountable at level $j < k$, and clinical deployment is
constrained by the accountable level, not the computed one.

\begin{figure}[htbp]
  \centering
  \includegraphics[width=0.75\textwidth]{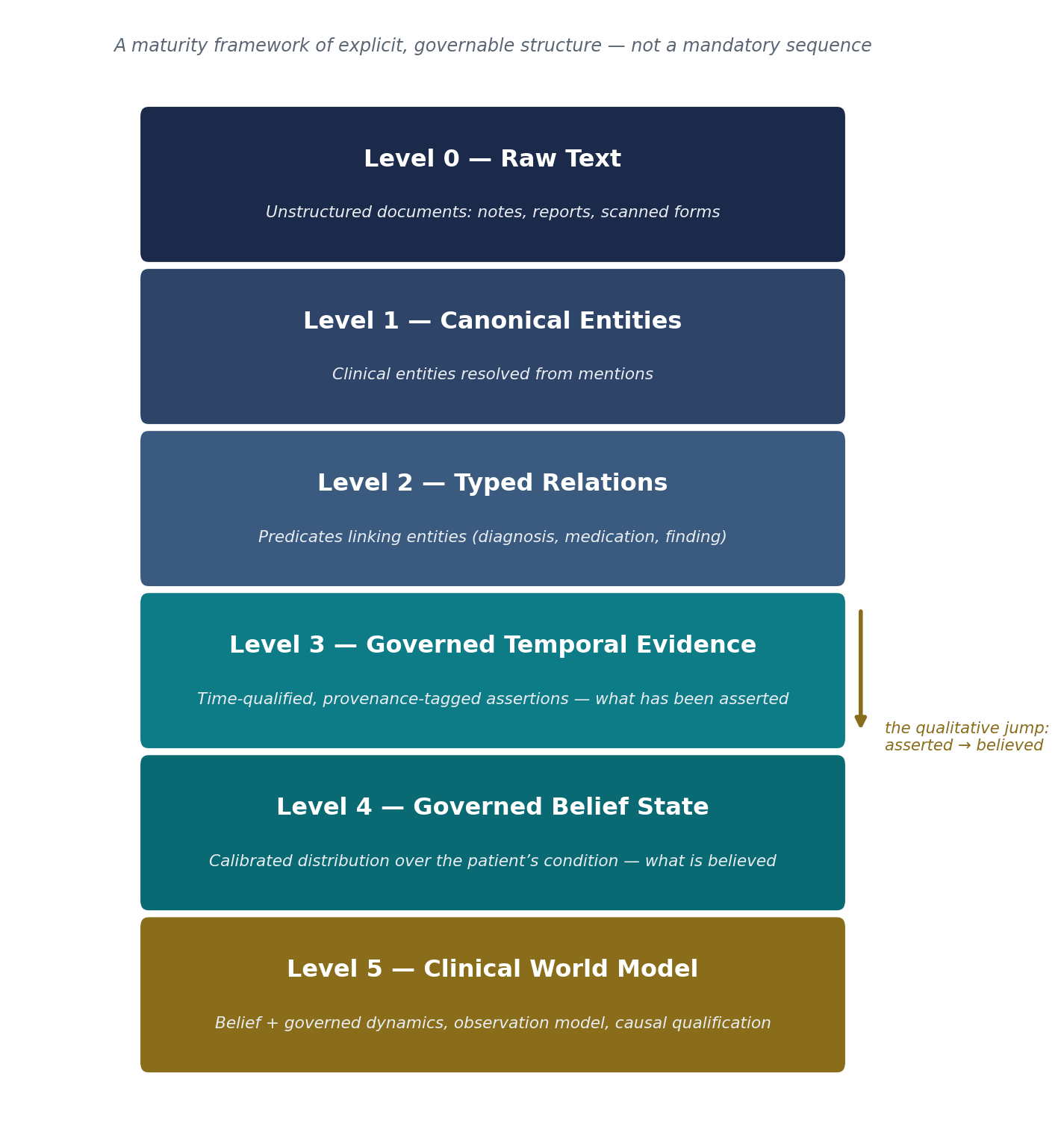}
  \caption{The six-level framework: increasing degrees of explicit, persistent,
  governable clinical structure. An architectural maturity framework, not a
  claimed law of intelligence.}
  \label{fig:maturity}
\end{figure}

With that guard in place, the framework of Figure~\ref{fig:maturity} orders
representations by how much clinical structure is explicit, persistent, and
governable: Level~0, raw text; Level~1, canonical entities; Level~2, typed
relations --- the classical clinical knowledge graph~\cite{hogan2021}; Level~3,
\emph{Governed Temporal Evidence} --- time-qualified, provenance-tagged
assertions, the temporal graph of what has been \emph{asserted}; Level~4, a
governed belief state over the patient's condition (the $\bel{x_t}$ of
equation~\eqref{eq:belief}); Level~5, a Clinical World Model --- belief plus
governed dynamics, an observation model, and causal qualification (future
work). In compact form: L0~Text $\to$ L1~Entities $\to$ L2~Relations $\to$
L3~Evidence $\to$ L4~Belief $\to$ L5~World Model. The count of levels is a
convention, not a claim --- a finer or coarser partition would serve --- and
what the framework actually asserts is one ordering of governance commitments
with a single qualitative jump: Level~3 to Level~4, from what has been
\emph{asserted} to what the system \emph{believes}. Levels~0--2 are a standard
natural-language-processing (NLP) and knowledge-graph pipeline; the jump to
belief is the step that record-as-state systems, however sophisticated their
predictors, do not expose. LLM-centric systems have high capability at low
maturity; that combination is exactly what
Section~\ref{sec:generated-vs-governed} diagnosed, and ``raising maturity to
meet capability'' is a fair one-sentence summary of this paper's program.

\section{Positioning, novelty, and what this paper does not claim}
\label{sec:positioning}

Several communities have built parts of what this paper assembles. Digital
twins in healthcare~\cite{katsoulakis2024} pursue persistent computational
models of individual patients, most maturely in mechanistic, organ- or
disease-specific form --- and we should not undersell them: a good mechanistic
twin already delivers individualized dynamics at higher fidelity than any
general belief state for the specific decision it was built for; the honest
claim is that our architecture is governance-general where the twin is
decision-specific, with the twin a natural high-fidelity dynamics component to
host inside the belief layer. Electronic-health-record (EHR) representation
learning (BEHRT and successors~\cite{li2020behrt}) learns predictive latent
representations over longitudinal coded records. The world-model literature
outside medicine --- latent dynamics models~\cite{hafner2023}, joint-embedding
predictive architectures~\cite{assran2023}, continuous-time
models~\cite{chen2018node}, and the belief-state formalism of partially
observable Markov decision processes (POMDPs)~\cite{kaelbling1998} --- supplies
the mathematical frame. Within medicine, EHRWorld~\cite{mu2026} demonstrates
that explicit temporally evolving state improves rollout stability over
frontier LLMs, and SMB-Structure (Standard Model
Biomedicine)~\cite{adam2026} --- whose title, ``The Patient is not a Moving
Document,'' shares our exact rhetorical move --- shows on large oncology and
pulmonary-embolism cohorts (23{,}319 and 19{,}402 patients respectively) that
joint-embedding predictive architecture (JEPA)-style embeddings capture disease
dynamics autoregressive baselines miss. A 2026 roadmap organizes biomedical
world models around data engines, simulators, and planning
substrates~\cite{wang2026}; and the term ``Clinical World Model'' itself is
used by Safavi-Naini et al.\ for a different purpose --- a
competency-evaluation framework~\cite{safavinaini2026} --- so we make no
priority claim over the term. Provenance and auditability are likewise not
novel in isolation: auditable, source-verified clinical-AI frameworks combining
RAG with provenance and tamper-evident logging exist~\cite{alu2026}, and
neurosymbolic clinical decision support with ontology grounding is an actively
worked area~\cite{garcez2009,besold2017}.

Three works from mid-2026 sharpen the neighborhood further.
MedBeads~\cite{nakajima2026} independently develops an immutable,
provenance-bearing clinical data substrate --- records as tamper-evident,
causally linked objects with amendments and retractions, supplied to agents by
deterministic traversal rather than semantic retrieval. It is closest to this
paper's Level-3 evidence layer, and it reinforces the claim that retrieval
alone is not a governance mechanism; our framework begins where such context
governance ends --- the governed evidence remains distinct from the canonical
belief state, the observation process, causal typing, and quarantined
simulation. Chen et al.'s review of medical world models~\cite{chen2026}
screened 1{,}455 records to find fourteen empirical systems, organizing the
field around state representation, dynamics, intervention-conditioned
simulation, and planning while naming causal identifiability, calibration,
uncertainty, and external validation as the field's open limitations --- a
capability-side map whose named gaps are, almost item for item, the
governance-side requirements this paper specifies. And CalTwin~\cite{khan2026}
directly targets calibration and distribution-shift robustness in latent
medical-world-model trajectories --- an example of the machinery that could one
day satisfy part of Definition~\ref{def:gcs}'s predictive tier, which is
precisely how the tiered standard is meant to be used.

Closest of all at the claim-typing layer is Qazi et al.'s ``Beyond Generative
AI''~\cite{qazi2025}, which grades clinical world models on a four-level
capability hierarchy --- temporal prediction, action-conditioned prediction,
counterfactual rollouts, and planning --- a rubric adjacent to the causal
typing this paper requires as artifact A4, and one any reader of this paper
should study alongside it. The distinction to keep sharp: theirs is an
evaluative hierarchy of what a model \emph{can do}; ours is enforced output
metadata --- every prediction carries its causal level, and promotion between
levels passes a per-query identification gate, not a capability judgment. The
``patient as evolving state, not a document'' thesis itself is likewise now
shared ground across the 2026 systems above, and we claim no priority over it.

Our novelty therefore cannot rest on ``neurosymbolic plus provenance plus
ontology,'' which is occupied, nor on the patient-as-state thesis, which is
shared. It rests on four commitments that remain under-occupied across all of
the works surveyed: (a)~awareness time as a first-class fourth temporal axis,
distinct from ingestion; (b)~the replay invariant --- belief as a deterministic
replay of the ledger prefix, making historical belief a computation rather than
an archaeology; (c)~quarantined simulation as a formal isolation construct
rather than a usage convention; and (d)~causal-level typing as enforced output
metadata with a per-query identification gate. These four, combined with the
belief-versus-evidence separation and the observation-process model into one
accountable operator, are the claim --- and the claim is a composition claim,
not a component claim. We do not assert novelty for the individual ingredients;
we propose that accountable longitudinal clinical reasoning requires their
governed composition around a replayable distinction between evidence and
belief. With $\Bstate$ the canonical stored belief-state object introduced in
Section~\ref{sec:five-objects}, the composition is
\[
  \Eaware \;\xrightarrow{\ \updop\ }\; \Bstate
  \qquad\text{and}\qquad
  (\Bstate, a) \;\xrightarrow{\ \text{simulation}\ }\; \Xbranch,
\]
with a governed, replayable update operator $\updop$ and no reverse write path
from simulation to evidence or belief. The architectural invariant, not any
component, is the contribution. One qualification keeps the replay commitment
honest: replay is deterministic with respect to the ledger prefix together with
the pinned update-operator version, model versions, ontology and rule versions,
configuration, and stored stochastic state where inference is sampled --- given
the same, replay reproduces the same governed belief state. This is precisely
why the predictive tier of Definition~\ref{def:gcs} requires model-version
binding; without it, the same ledger could yield a different posterior after a
model update, and the invariant would silently fail. That is a synthesis claim,
not a claim to have invented a category; and it is an architectural claim whose
empirical standing depends entirely on the research program sketched in the
conclusion.

\section{Conclusion}
\label{sec:conclusion}

This paper set out to relocate the debate about clinical AI: from what language
models can generate to what clinical systems must maintain. Its argument,
compressed: clinical reasoning over a patient record is state estimation under
partial observability; the interface of an autoregressive language model,
whatever its internal representations, exposes no governed state to estimate
with; and the remedy is architectural. The accountability demand decomposes
into four information requirements --- an evidence ledger with awareness-time
versioning, a belief state distinct from the record, an observation model, and
causal typing of predictive claims --- and the maturity framework gives the
field a way to say precisely what a system makes governable, as opposed to what
it can compute. Fluency is now abundant; governed clinical state is not --- and
it is the difference between a system that talks about patients and a system
that can be held to account for what it believes about them.

\paragraph{The research questions, answered as far as this paper can.}
The introduction posed four questions; here is the state of each.

\begin{itemize}
  \item \textbf{RQ1 (temporal consistency).} Answer at this stage: not yet
  empirically answered --- predicted by the architecture. The architectural
  analysis predicts improved temporal consistency and provenance, because
  governed state exposes information that retrieval-only architectures do not
  guarantee; the contrast of Section~\ref{sec:one-question} and the constructed
  longitudinal case of Section~\ref{sec:longitudinal-case} demonstrate the
  mechanism, not the margin. The quantitative comparison against strong
  long-context and retrieval baselines remains to be measured, and is the first
  measurement to be taken up in future work.

  \item \textbf{RQ2 (multi-horizon dynamics).} Answer at this stage: not yet
  answerable --- and this paper says precisely why. The maturity framework
  locates what the question tests (the Level-3 to Level-4 jump), and answering
  it requires an operational belief state first; that operationalization is the
  critical path of the program.

  \item \textbf{RQ3 (verification).} Answer at this stage: the question is now
  well-posed. Definition~\ref{def:gcs} (Governed Clinical State) supplies the
  constraint classes a verifier enforces; the experiment is cheap to run and
  informative either way, including against us.

  \item \textbf{RQ4 (transportability).} Answer at this stage: predicted, not
  demonstrated. The observation-model artifact (A3) of the accountability
  decomposition predicts the transport benefit; before any measurement can be
  trusted, future work must state the estimation procedure and its
  identifiability assumptions.
\end{itemize}

Each question will be taken up in future work with quantitative thresholds
preregistered before results are reported --- and each can go against us: a
reader who believes that scale, context length, or retrieval make governed
state unnecessary will find in these four questions the concrete places to show
it.

\paragraph{Future work.}
Two fronts, in order. \emph{Empirical}: answer RQ1--RQ4, beginning with RQ1 and
RQ3 (which need only the evidence layer already in production plus a minimal
belief layer), then RQ2 and RQ4 (which require the operational belief state and
observation model first). \emph{Specification}: publish the full engineering
treatment of the evidence ledger and governed update operator; an operational
schema for the belief state; the observation model's estimation procedure with
its identifiability assumptions stated (fitting a measurement policy
conditioned on latent state requires the very state being estimated); and the
per-query causal identification contracts that govern promotion between
prediction levels.

\paragraph{About MyndwareMed.}
This paper is the conceptual foundation of a system being built. MyndwareMed's
platform implements the governed evidence layer specified here ---
ontology-bound, provenance-tagged, multitemporal --- in production today
(maturity Level~3), is engineering the governed belief layer (Level~4), and is
researching the full Clinical World Model (Level~5) against the research
questions above. The architecture in this paper is, deliberately, the standard
that program is accountable to.

\paragraph{Competing interests and scope statement.}
Both authors are employees of MyndwareMed, which develops a commercial platform
implementing the architecture this paper specifies; the reader should weigh the
positioning claims of Section~\ref{sec:positioning} accordingly. This paper is
a position-and-specification paper: it uses no patient data, reports no
experiments, and claims no empirical or clinical result. Statements about
MyndwareMed's implementation are claims about engineering artifacts, not about
demonstrated clinical performance, and the commitments outlined in the
conclusion will be preregistered with quantitative thresholds before any
results are reported.

%==============================================================================
%  References
%
%  The original document uses a hand-maintained, Nature-style reference list
%  whose numbering is fixed by the order of the \bibitem entries below (it is
%  NOT the order produced by any standard .bst). Keep this order if you want
%  the citation numbers of the published PDF to be reproduced exactly.
%  A BibTeX version is provided in references.bib for convenience.
%==============================================================================

\end{document}